\documentclass[a4paper]{article}
\usepackage{INTERSPEECH_v2}
\usepackage{hyperref}
\usepackage{url}
\usepackage{amsmath,amssymb,amsthm}
\usepackage{color}
\usepackage{graphicx}
\usepackage{subcaption}
\usepackage{floatrow}
\newfloatcommand{capbtabbox}{table}[][\FBwidth]
\usepackage{blindtext}
\usepackage{psfrag}
\usepackage{epstopdf}
\usepackage{floatrow}
\newcommand{\h}{\ensuremath{\mathbf{h}}}

\newcommand{\x}{\ensuremath{\mathbf{x}}}
\newcommand{\y}{\ensuremath{\mathbf{y}}}
\newcommand{\z}{\ensuremath{\mathbf{z}}}

\newcommand{\bphi}{\ensuremath{\boldsymbol{\phi}}}

\newcommand{\btheta}{\ensuremath{\boldsymbol{\theta}}}

\newcommand{\bbE}{\ensuremath{\mathbb{E}}}

\newcommand{\calN}{\ensuremath{\mathcal{N}}}

{%
\begin{list}{#1}{
\vspace{-\topsep}
\vspace{-\partopsep}
\setlength{\itemindent}{0cm}
\setlength{\rightmargin}{0cm}
\setlength{\listparindent}{0cm}
\settowidth{\labelwidth}{#1}
\setlength{\leftmargin}{\labelwidth}
\addtolength{\leftmargin}{\labelsep}
\setlength{\itemsep}{0cm}
}%
}%
{%
\end{list}
\vspace{-\topsep}
\vspace{-\partopsep}
}

{\begin{enumerate}%
}%
{\end{enumerate}}

\theoremstyle{plain}% default

\newtheorem*{prop*}{Proposition}

\theoremstyle{definition}

\newtheorem*{defn*}{Definition}

\newtheorem*{exmp*}{Example}

\newtheorem*{conj*}{Conjecture}

\theoremstyle{remark}

\newtheorem*{rmk*}{Remark}

\title{Acoustic Feature Learning \\ via Deep Variational Canonical Correlation Analysis}
\name{Qingming Tang$^1$, Weiran Wang$^1$, Karen Livescu$^1$}
\address{
  $^1$Toyota Technological Institute at Chicago, USA}
\email{\{qmtang,weiranwang,klivescu\}@ttic.edu}

\begin{document}

\maketitle
\begin{abstract}
We study the problem of acoustic feature learning in the setting where we have access to another (non-acoustic) modality for feature learning but not at test time.  We use deep variational canonical correlation analysis (VCCA), a recently proposed deep generative method for multi-view representation learning.  We also extend VCCA with improved latent variable priors and with adversarial learning. Compared to other techniques for multi-view feature learning, VCCA's advantages include an intuitive latent variable interpretation and a variational lower bound objective that can be trained end-to-end efficiently. We compare VCCA and its extensions with previous feature learning methods on the University of Wisconsin X-ray Microbeam Database, and show that VCCA-based feature learning improves over previous methods for speaker-independent phonetic recognition.

\end{abstract}

\section{Introduction}
\label{sec:intro}

Applications involve acoustic speech often benefit from one or more additional types of measurements (e.g. images, video and articulatory measurements) via representing and reasoning about the multiple modalities in the same vector space ~\cite{hazen+etal_interspeech07,chrupala2017representations, harwath2016unsupervised, huang2013audio, arora2013multi,badino2016integrating}. A number of approaches have been developed to automatically learn latent spaces that are modality-agnostic. In this work we address the setting where multiple modalities, or ``views'', are available for feature learning but only one (in this case, acoustics) is available for downstream tasks (e.g., speech recognition). 
A popular class of methods in this area is based on canonical correlation analysis (CCA,~\cite{hotelling1936relations}) and its nonlinear extensions~\cite{melzer2001nonlinear,bach2002kernel,andrew2013deep}. CCA finds projections of two data views that are maximally correlated, subject to uncorrelatedness between learned dimensions in each projected view. This common subspace or projection mappings are often more discriminative or robust to noise, and can be used as good features for downstream tasks.

In this paper, we explore a recently proposed deep generative variant of CCA, deep variational CCA (VCCA)~\cite{wang2016deep}, for multi-view acoustic feature learning.  VCCA models the joint distribution of two views with a latent variable model capturing both the shared and private (view-specific) information, where the distributions are parameterized by deep neural networks (DNNs).  VCCA optimizes a variational lower bound of the 
% data
likelihood and allows for straightforward training using small minibatches of training samples. 
VCCA has been found to improve over % deep CCA and 
other multi-view methods for multiple real-world tasks, but thus far not for speech tasks~\cite{wang2016deep}.  In this work we study and extend VCCA and show that it can learn acoustic features from parallel acoustic and articulatory data that improve phonetic recognizers.

We compare VCCA and its extensions with previous single-view and multi-view feature learning approaches using the University of Wisconsin X-ray Microbeam Database~\cite{westbury1990x}, which consists of simultaneously recorded audio and articulatory measurements.  We first show that VCCA learns better acoustic features for speaker-independent phonetic recognition.
%, for different sizes of input context window.
We then extend VCCA in several ways, including aspects of the prior distributions and adversarial learning~\cite{goodfellow2014generative}.

\section{Deep variational CCA}
\label{sec:vcca}

\begin{figure*}[t]
  \centering
  \begin{tabular}{@{}c@{\hspace{0.05\linewidth}}|@{\hspace{0.04\linewidth}}c@{}}
    \psfrag{x}[][]{$\x$}
    \psfrag{y}[][]{$\y$}
    \psfrag{z}[][]{$\z$}
    \psfrag{hx}[][]{$\h_x$}
    \psfrag{hy}[][]{$\h_y$}
    \raisebox{1.5ex}{\includegraphics[width=0.20\linewidth]{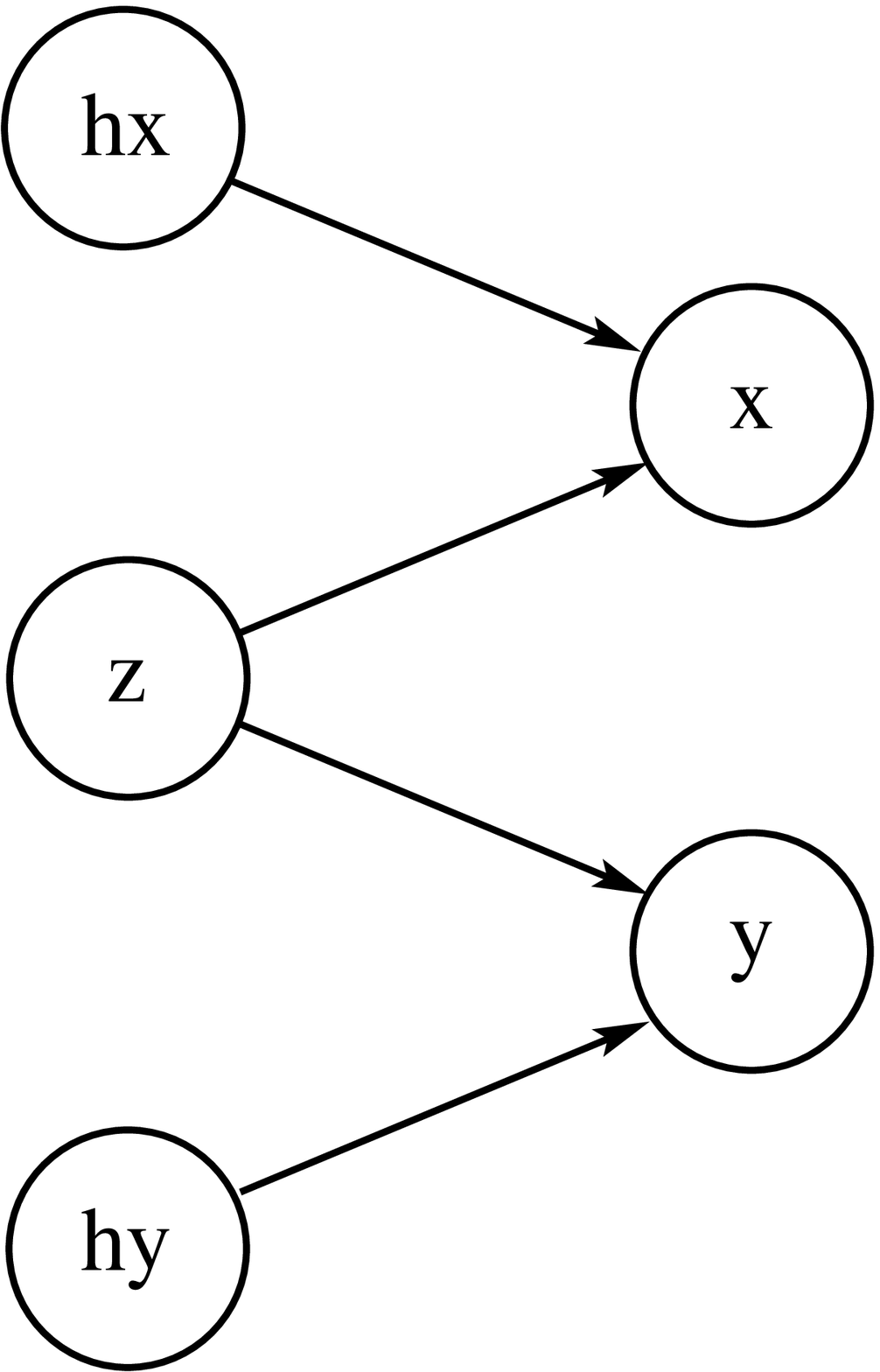}} & 
    \psfrag{x}[][][1.0]{$\x$}
    \psfrag{y}[][][1.0]{$\y$}
    \psfrag{z}[][][1.0]{$\z$}
    \psfrag{q(hx|x)}[][][0.9]{$q_{\bphi} (\h_x | \x)$}
    \psfrag{q(z|x)}[][][0.9]{$q_{\bphi} (\z | \x)$}
    \psfrag{q(hy|y)}[][][0.9]{$q_{\bphi} (\h_y | \y)$}
    \psfrag{p(z)}[][][1.0]{$p(\z)$} %{$\z = \bmu + \bSigma \bepsilon$ }
    \psfrag{p(hx)}[][][1.0]{$p(\h_x)$}%{$\h_x = \bmu + \bSigma \bepsilon$ }
    \psfrag{p(hy)}[][][1.0]{$p(\h_y)$}%{$\h_y = \bmu + \bSigma \bepsilon$ }
    \psfrag{p(x|z)}[][r][0.8]{$p_{\btheta} (\x | \z, \h_x)$}
    \psfrag{p(y|z)}[][r][0.8]{$p_{\btheta} (\y | \z, \h_y)$}
    \psfrag{D1}[l][][1.0]{$D_1$}
    \psfrag{D2}[l][][1.0]{$D_2$}
    \includegraphics[width=0.56\linewidth]{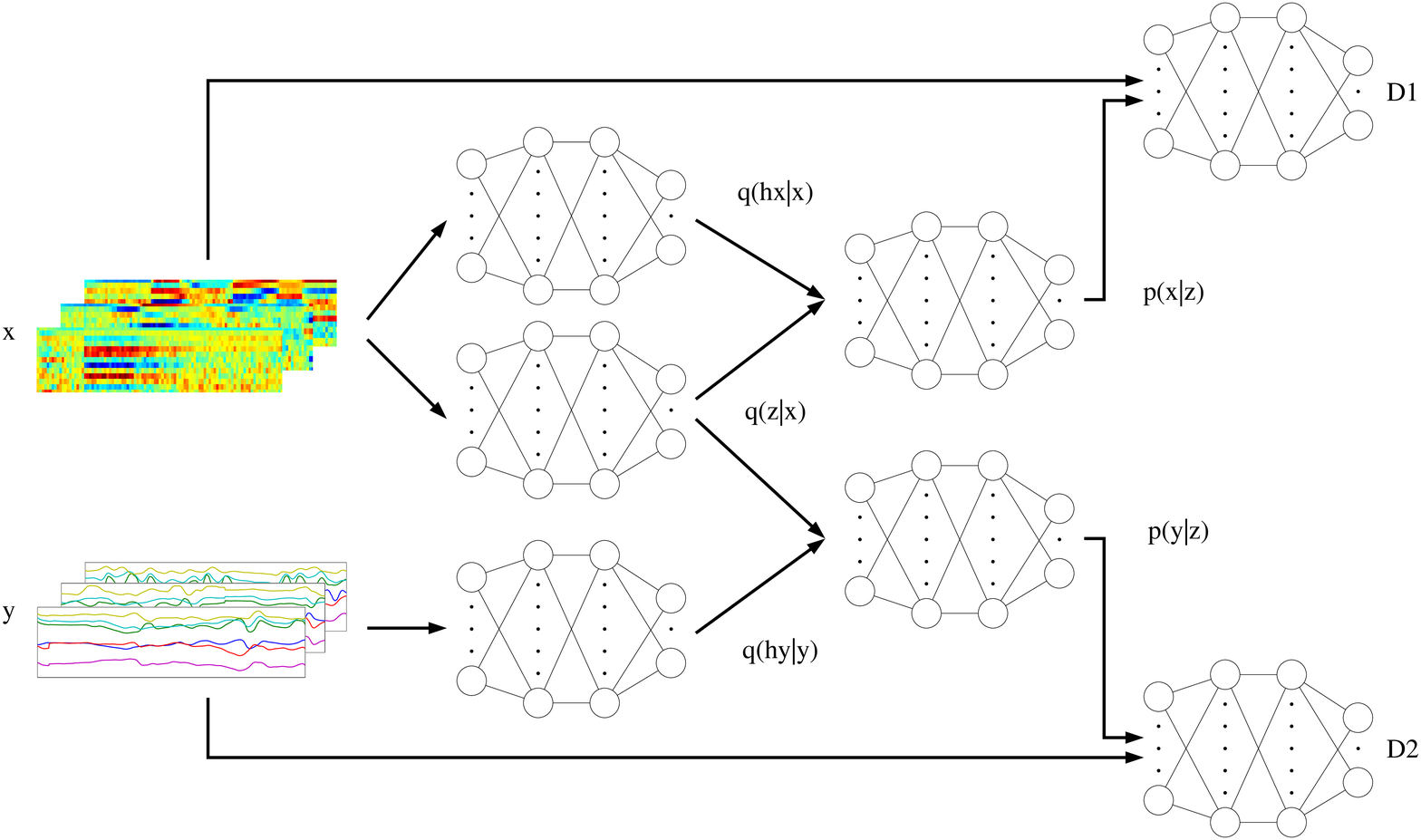} 
  \end{tabular}
  \caption{Left: graphical model of  VCCA with view-specific private variables (VCCAP). Right: multi-view acoustic feature learning with VCCAP and adversarial training.  VCCAP alone corresponds to this model without D1, D2; basic VCCA also has no $h_x, h_y$.}
  \label{fig:vcca-private}
\end{figure*}

Deep CCA (DCCA,~\cite{andrew2013deep}), which uses DNNs for projecting two views into a common subspace, has achieved excellent empirical performance for tasks across several domains in the setting of unsupervised multi-view feature learning~\cite{wang2015unsupervised, yan2015deep, lu2015deep, wang2015deep}. However, DCCA has certain drawbacks.  First, its objective couples all training samples together and is hard to optimize with stochastic gradient descent (SGD). (Although this is somewhat alleviated by~\cite{Wang_15c}, their optimization algorithm requires more tuning parameters.) 
Second, DCCA focuses on extracting the shared information (most correlated dimensions) only, while there may also be useful view-specific (``private'') information that we wish to retain. An autoencoder-based extension to DCCA has been proposed~\cite{wang2015deep}, but in practice it produces quite similar results to DCCA. Finally, DCCA (and kernel CCA) are deterministic methods, that is they do not provide a probabilistic model of the data, so they can not generate samples of the input modalities.\footnote{This final point is not relevant to our experimental setting in this paper, but can be important more broadly in other settings.}

Deep variational CCA is based on extending a latent variable formulation of CCA due to Bach and Jordan~\cite{bach2005probabilistic}.
%established the equivalence between linear CCA and a latent variable model as follows.
Denote the two input views $x$ and $y$, and assume they are both generated independently from a common (multi-dimensional) latent variable $z$:  $p(x,y,z)=p(z) p(x|z) p(y|z)$. Bach and Jordan~\cite{bach2005probabilistic} showed that, when $p(z)$, $p(x|z)$, and $p(y|z)$ are Gaussian, with the conditional means linear in $z$, the maximum likelihood solution gives ``projections'' $\bbE[z|x]$ and $\bbE[z|y]$ that lie in the same space as the (deterministic) linear CCA projections. 
%We note that t
The intuition of a common variable generating multiple views fits our setting, as in our case there is a latent variable, the phonetic transcription, that generates (to a large degree) both acoustic and articulatory measurements and that we wish to recover. However, modeling the complex distribution of speech is too challenging for linear mappings; we therefore consider a deep extension of this generative model.
 
\subsection{Basic variational CCA}

VCCA was introduced in~\cite{wang2016deep}; we review the formulation here briefly. In VCCA, the prior $p(z)$ is taken to be a simple Gaussian, but $p_{\theta} (x|z)$ and $p_{\theta} (y|z)$ are isotropic Gaussians with means parameterized by two DNNs respectively (parameters collectively denoted $\theta$) and whose standard deviations are a tuning parameter. With this parameterization, the marginal distribution $p(x,y)=\int p(z) p(x|z) p(y|z) dz$ does not have a closed form. Using a variational inference approach similar to that of variational autoencoders~\cite{kingma2014auto}, VCCA parameterizes the approximate posterior $q_{\phi} (z|x)$ using another DNN (with weight parameters $\phi$), and optimizes the following lower bound of the log likelihood: 
\begin{align} \label{eqn:likelihood}
\mathcal{L}(x,y;\theta,\phi) & = -D_{KL}(q_{\phi}(z|x)||p(z)) \nonumber \\ 
& \quad + E_{q_{\phi}(z|x)}[\log{p_{\theta}(x|z)}+\log{p_{\theta}(y|z)}] .
%%-D_{KL}(q_{\phi}(z|x)||p(z)) + E_{q_{\phi}(z|x)}[\log{p_{\theta}(x|z)}+\log{p_{\theta}(y|z)}] .
\end{align}
Here $D_{KL}(q_{\phi}(z|x)||p(z))$ denotes the KL divergence between the learned $q_{\phi}(z|x)$ and the prior $p(z)$. Note that, for Gaussian observation models $p(x|z)$ and $p(y|z)$, maximizing the likelihood is equivalent to minimizing the reconstruction errors. Calculating the expectation term in~\ref{eqn:likelihood} can be done efficiently by Monte Carlo sampling. 
The approximate posterior is modeled as another Gaussian whose mean and standard deviation are outputs of the neural network with weights $\phi$. 
Using the ``reparameterization trick'' of~\cite{kingma2014auto}, we only need to draw samples from the simple $\calN(0,I)$ distribution to obtain samples of $q_{\phi} (z|x)$, and stochastic gradients with respect to the DNN weights can be computed using standard backpropagation.

Given a set of $N$ paired samples $\{(x_1,y_1),...,(x_N,y_N)\}$ of the input variables $(x,y)$, VCCA  maximizes the variational lower bound on the training set 
%\vspace{-.1in}
\begin{equation*}
\max_{\theta,\phi}\; \sum_{i=1}^N {\mathcal{L}(x_i,y_i;\theta,\phi)}.
\end{equation*}

Finally, as in the probabilistic interpretation of CCA, we use the mean of $q_{\phi} (z|x)$ (the outputs of the corresponding DNN when fed $x$ as input) as the learned features.

\subsection{VCCA with private variables  (VCCAP)}

VCCA has been extended by~\cite{wang2016deep} to include view-specific private variables; we denote the resulting model VCCA-private (VCCAP). 
VCCAP introduces two additional sets of latent variables, $h_x$ of dimensionality $d_{h_x}$ and $h_y$ of dimensionality $d_{h_y}$, for the two views $x$ and $y$ respectively.  The input $x$ is taken to be jointly generated by $h_x$ and $z$, and similarly for the second view.  The intuition is that there might be
%large variations of the inputs
aspects of the two views that are not captured by the shared variables, so that reconstructing the inputs using only the shared variables is difficult.  For example, in the case of acoustic and articulatory data used in our experiments, the articulatory data does not contain nasality information while the acoustic signal does.  Explicitly modeling the private information in each view eases the burden of the reconstruction mappings, and may also retain additional useful information in the target view. The graphical model of VCCAP is shown in Figure \ref{fig:vcca-private} (left). 

To obtain a tractable objective, VCCAP parameterizes two additional posteriors $q_{\phi} (h_x|x)$ and $q_{\phi} (h_y|y)$ and maximizes the following lower bound on the log likelihood:
\begin{gather}
\mathcal{L}_{private}(x,y;\theta,\phi) = - D_{KL}(q_{\phi}(z|x)||p(z))\nonumber \\
- D_{KL}(q_{\phi}(h_y|y)||p(h_y)) - D_{KL}(q_{\phi}(h_x|x)||p(h_x)) \nonumber \\
+ E_{q_{\phi}(z|x)q_{\phi}(h_x|x)}[\log{p_{\theta}(x|z,h_x)}] \nonumber \\
+ E_{q_{\phi}(z|x)q_{\phi}(h_y|y)}[\log{p_{\theta}(y|z,h_y)}]
\label{eqn:vcca-private}
\end{gather}
Here we use simple $\calN(0,I)$ priors for $h_x$ and $h_y$. 
Given a training set, we maximize this variational lower bound on the training samples using SGD and Monte Carlo sampling as before. Note that, unlike DCCA, the VCCA and VCCAP objectives are sums over training samples, so it is trivial to use SGD with small minibatches. Again, we use the mean of $q_{\phi} (z|x)$ as the learned features for VCCAP.

\section{Extensions of VCCA}
\label{sec:extension}

We now propose two extensions of VCCA/VCCAP, which will be explored later in our experiments. 

\subsection{Tuning the KL divergence terms}
\label{sec:tuning_kl}

The KL divergence terms in the objectives of VCCA and VCCA-private naturally arise in the variational lower bound derivation. On the other hand, they can also be understood  as regularization terms that control the complexity of the projection networks $q_{\phi}(z|x)$, $q_{\phi}(h_x|x)$ and $q_{\phi}(h_y|y)$: They enforce our belief that the posterior distribution should be close to $\calN(0, I)$ so that each dimension has approximately zero mean and unit scale, and different dimensions are independent.

Taking this regularization view, we can further control the regularization effect in two ways. First, we can add a weight $\beta$ on the KL divergence terms ($\beta=1$ for standard VCCA/VCCAP). Second, we could add more informative priors than $\calN(0,I)$. For our ASR task, we typically concatenate input frames over a context window. We find that features learned at a moderate window size can be used as priors for learning at a larger context window size. Such priors may both improve the feature quality and speed up training over using the simple Gaussian prior. Experiments demonstrating this effect are given in Section~\ref{sec:kl-result}.

\subsection{Generative adversarial training}
\label{sec:gan}

Adversarial learning is an increasingly popular approach for learning deep generative models~\cite{goodfellow2014generative, donahue2017adversarial, dumoulin2016adversarially, chen2016multi, arjovsky2017wasserstein, zhao2017energy}. In this approach there are two modules: a discriminator that takes a data sample and tries determine whether it is generated by a model (fake) or comes from the real data distribution (e.g., a sample from the training set), and a generator that produces data samples from the generative model and tries to fool the discriminator. At equilibrium, generative adversarial networks (GANs) should be able to capture well the data distribution. 

We extend VCCA/VCCAP with generative adversarial training by viewing the basic VCCA/VCCAP model as a pair of generators, and adding two discriminators $D_1$ and $D_2$, one for each view, which try to distinguish the generated samples from training set input samples. Accordingly, we augment the original VCCA objective with losses on the discriminators. For fixed generators, $D_1$ optimizes the following objective for the $i$-th sample ($D_2$ maximizes an analogous objective):
\begin{eqnarray}
\max_{D_1}\; \log{D_1(x_i)} + \log(1-D_1(x'_i)) 
\label{eqn:discriminators}
\end{eqnarray}
where $x_i$ is the original input of the first view and $x'_i$ is the corresponding reconstruction. %$D_1$ tries to output $1$ for true samples and $0$ for reconstructions.
Given the discriminators, the generators (including the projection networks and reconstruction networks) optimize the following objective for the $i$-th sample
\begin{gather*}
\max_{\theta,\phi}\; \mathcal{L}_{private}(x_i,y_i;\theta,\phi) \\
+ \lambda_1 \log (D_1 (x'_i))
+ \lambda_2 \log (D_2 (y'_i)).
\label{eqn:regularized generator}
\end{gather*}
Here $\lambda_1$ and $\lambda_2$ are two hyperparameters used to trade off the GAN loss with the VCCA/VCCAP loss. Note that $x'_i$ and $y'_i$ are functions of the generators (and thus also parameterized by $\theta,\phi$), and the generators try to make $D_1$ and $D_2$ accept fake samples (reconstructions). 
We alternate the above two stages during training. The model architecture incorporating GANs is depicted in Figure~\ref{fig:vcca-private} (right). Experiments demonstrating this approach are given in Section~\ref{sec:gan-result}.
\section{Experimental Results}
\label{sec:experiment}

The Wisconsin X-ray microbeam (XRMB) \cite{westbury1990x} corpus consists of both speech and articulatory measurements, which are simultaneously recorded, from 47 American English speakers. Previous works ~\cite{arora2013multi,wang2015unsupervised} have shown that features learned by linear, kernel and deep CCA can be used to improve performance of speaker-independent phonetic recognition on this corpus. Note that the recognizer is trained and tested using acoustic view only though features are learned using both views.

Following the basic setup of~\cite{wang2015unsupervised}, the two input views are based on standard $39D$ MFCC+$\Delta+\Delta\Delta$s and $16D$ articulatory features (horizontal/vertical displacement of 8 pellets attached to several parts of the vocal tract). To incorporate context information, the inputs are concatenated over a $W$-frame window centered at each frame, giving $39\times W$ and $16\times W$ feature dimensions for each of the two views respectively.

The $47$ speakers of XRMB are divided into disjoint sets of $35/12$ speakers for feature learning and recognition respectively. The $12$ recognition speakers are further divided into 6 disjoint groups to perform a $6$-fold experiment. In each fold, $8/2/2$ speakers are used for training/tuning/testing the recognizer respectively. For each fold, we choose the hyperparameters that give the best PER on the fold-specific dev set and measure its corresponding test result; finally, we report the average test PER over the 6 folds. Each split/fold is gender-balanced, and the per-speaker mean and variance of the articulatory measurements are removed for the feature learning stage. Unlike~\cite{wang2015unsupervised}, we include silence frames in feature learning, yielding a total of 1.7M training samples. (Although the added frames correspond to silence, the context window often overlaps speech.) This larger training set benefits the generative VCCA/VCCAP methods.

\subsection{Algorithms and architectures}
We compare VCCA/VCCAP and its extensions with baseline MFCCs, multi-view features learned with deep CCA (DCCA~\cite{andrew2013deep,wang2015unsupervised}) and multi-view contrastive loss (Contrastive~\cite{hermann2014multilingual}). The contrastive loss aims to make the distance between paired samples smaller than the distance between unpaired (negative) samples by a margin. Note that in previous work~\cite{wang2015unsupervised}, DCCA was already found to outperform supervised DNN bottleneck features (trained on the recognizer training data and used in a tandem recognizer~\cite{hermansky2000tandem}) for $W=7$. Both DCCA and Contrastive were compared against VCCA/VCCAP in~\cite{wang2016deep} for a $7$-frame context window, where VCCAP was slightly outperformed by Contrastive when silence frames are not used in training. In this paper, we show that VCCAP and its extensions outperform these alternatives when all methods use the same larger training set at different context sizes.

DNN architectures and training procedures for DCCA and Contrastive used here are consistent with previous work~\cite{wang2016deep}, and the VCCA-based methods use the same architectures. Each DNN has up to $3$ ReLU~\cite{nair2010rectified} hidden layers: each hidden layer of $q(z|x)$ has $1500$ units while each hidden layer of $q(h_x|x)$ and $q(h_y|y)$ has $1024$ units, and the discriminators $D_1$ and $D_2$ use $[2048, 1500, 1500]$ units in each hidden layer. With limited tuning at context size $W=7$, we fix the standard deviations of %the Gaussian distributions
$p_{\theta}(x|z,h_x)$ and $p_{\theta}(y|z,h_y)$ to $1$ and $0.1$ respectively.
Our methods are trained using the Adam optimizer~\cite{kingma2015adam} with a fixed learning rate of $0.0001$ and minibatch size $200$, except that the discriminators of VCCAP+GAN are updated less frequently using minibatches of size $1800$. Dropout~\cite{srivastava2014dropout} at a rate of $0.2$ is used for all layers. The overall best performing VCCA, VCCAP and DCCA models have a feature dimensionality of $70$, while the best feature dimensionality is $50$ for Contrastive. We train the four models to convergence (around $60$, $300$, $60$ and $30$ epochs for VCCA, VCCAP, DCCA and Contrastive loss, respectively).

We use two phone recognizers in this paper. The first is a 3-state monophone HMM/GMM system, which is fast and consistent with previous work~\cite{wang2015unsupervised}.  This recognizer is built with Kaldi~\cite{povey2011kaldi} and uses a TIMIT bigram language model. 
The learned features are used in a tandem approach~\cite{hermansky2000tandem} for this recognizer, i.e., they are concatenated with the original $39D$ MFCCs and used as input to the HMM/GMM recognizer. 
% This recognizer uses a TIMIT bigram language model. 
The second recognizer is a connectionist temporal classification (CTC)-based recognizer~\cite{graves2006connectionist}, built with TensorFlow~\cite{abadi2016tensorflow}, that uses a $2$-layer bidirectional recurrent neural network (RNN) with $512$ gated recurrent unit (GRU~\cite{chung2014empirical}) cells in each layer. We do not concatenate features with the original MFCCs as this tends to give better performance on dev sets.  No language model is used for this recognizer. We decode with beam search with a beam width of $100$. Since experiments with the CTC-based recognizer are much slower than with the HMM/GMM, we provide CTC % recognition
results for a subset of the experiments.

\begin{figure}[t]
  \centering
  \includegraphics[width=0.96\textwidth]{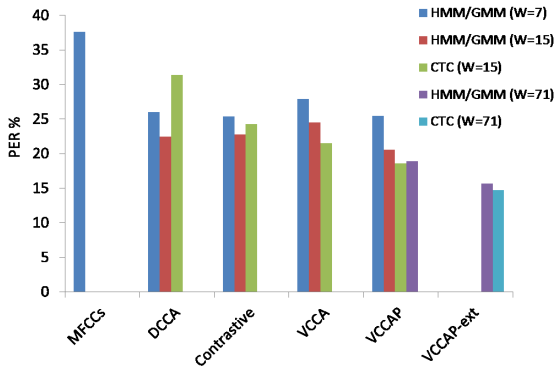}
%  \vspace{-.1in}
  \caption{Average test set phone error rate (PER, \%) over $6$ folds obtained by HMM/GMM and CTC-based recognizers with features learned by different methods at window size $W=7/15/71$.  VCCAP-ext is the extension of VCCAP with a weighted KL divergence term and learned priors.}
  \label{fig:demon_vccap}
\end{figure}

\vspace{-.05in}
\subsection{Main test set results and effect of window size}
\label{sec:comvccap}

We first give the main final test set results, summarized in Figure~\ref{fig:demon_vccap}, and then describe additional comparisons on the dev set.
We observe from Figure~\ref{fig:demon_vccap} that all deep multi-view features benefit from wider context, and significantly improve over the original MFCC features. Using either HMM/GMM or CTC-based recognizers, VCCAP outperforms basic VCCA, is comparable to Contrastive and DCCA for $W=7$, and outperforms all alternative methods for $W=15$.  We therefore focus on VCCAP in the remaining experiments and discussion.  The best performance is obtained with the extension to VCCAP proposed in Section~\ref{sec:tuning_kl}, with a weight on the KL divergence term (of $\beta = 10$) and where the $\calN(0,I)$ prior for a large-window input (here $W=71$) is replaced by the posterior learned with a smaller window (here $W=35$).  

Table~\ref{tab:window_size} provides a more detailed look at the role of the context window size, in terms of development set PER in one fold (to avoid excessive test set reuse).  When fixing the other hyperparameters of VCCAP, the PER initially decreases quickly with increasing context window size, then plateaus after $W = 15$, and starts to increase again at $71$ frames.  It is only by using the learned prior extension that we are able to obtain further improvements with larger context windows.

\begin{table}[t]
  \centering
  \begin{tabular}{|c||c|c|c|c|c|c|c|}
\hline
$W$ & 3 & 7 & 15 & 35 & 41 & 61 & 71 \\
\hline\hline
PER & 27.4 & 22.8 & 19.5 & 15.1 & 15.6 & 14.9 & 16.1 \\
\hline
  \end{tabular}
    \caption{Development set PER (\%) of one fold obtained with the HMM/GMM recognizer and VCCAP features for different input context sizes when trained for $60$ epochs.} 
  \label{tab:window_size}
\end{table}

\vspace{-.05in}
\subsection{VCCAP Extensions}
We next provide a more detailed study of the proposed VCCAP extensions. To avoid excessive test set reuse and to speed up experimentation, we report dev set results from a single fold when each model is trained for $60$ epochs. We then use the learned features only in the HMM/GMM recognizer.

\begin{table}[t]
  \caption{Development set PER (\%) 
obtained with the HMM/GMM recognizer and VCCAP features for different KL divergence term penalty weights, with/without priors learned by the 35-frame window model.}
  \label{tab:posterior}
  \centering
  \begin{tabular}{|l||c|c|c|c|}
    \hline
   & $W=35$ & 51 & 61 & 71 \\
	\hline\hline

 $\beta=1$, $\calN(0,I)$ prior & 15.1 & 14.9 & 14.9 & 16.1 \\
	\hline
 $\beta=10$, $\calN(0,I)$ prior & 15.1 & 14.2 & 15.1 & 14.7 \\
  	\hline
 $\beta=10$, learned prior & - & 14.4 & 13.8 & 13.2 \\
	\hline
  \end{tabular}
\end{table}

\subsubsection{Tuning the KL divergence terms}
\label{sec:kl-result}
Table~\ref{tab:window_size} shows that learning from very large context windows is more challenging.
Table~\ref{tab:posterior} shows that the proposed VCCAP extensions of Section~\ref{sec:tuning_kl} allow for further gains with larger windows.  Specifically, with a very large window of $W=71$, giving more weight to the KL divergence term improves performance, and using a prior corresponding to the learned posterior from the $W=35$ model helps further.  These extensions force the model to pay more attention on the central $35$ frames.

\subsubsection{Incoporating adversarial learning}
\label{sec:gan-result}

Finally, we experiment with the VCCAP+GAN approach (Section~\ref{sec:gan}). We train VCCAP with the best hyperparameters for $30$ epochs; then starting from this model we train VCCAP+GAN and VCCAP both for another $30$ epochs each. Tradeoff parameters for the discriminators ($\lambda_1=\lambda_2=5$) are tuned at $W=7$ based on average dev set PER. VCCAP+GAN slightly outperforms VCCAP, with gains of $0.1$-$0.4$ in dev set PER across folds for $W=7/15/35$, but the improvement is less stable than that of the other extensions. Considering the added complexity of GAN training, the results may not justify adopting this approach in general, and for our final test results we did not include this extension.

%% \input{related}
%\vspace{-.05in}
\section{Conclusion}
\label{sec:conclusion}

We have shown that VCCA-private and its extensions improve over previous approaches for unsupervised acoustic feature learning from multi-view acoustic and articulatory data, and in particular that the method benefits from large input context windows using learned priors.  In fact, we have observed that we can continue to improve performance with even larger windows by re-learning priors, but thorough experimentation becomes impractical; this suggests that recurrent models may be fruitful, as a way of handling very large context.
Further avenues for research 
in unsupervised speech processing
include using the extracted private variables (which are unused in our work thus far), which may be useful for generation tasks like voice morphing and other synthetic speech generation.  The setting could also be extended to semi-supervised learning, that is joint training of the recognizer and VCCA objectives, as well as for learning sequence-level features.

%\vspace{-.05in}
\section{Acknowledgments}
This research was supported by NSF grant IIS-1321015 and used GPUs donated by NVIDIA Corporation. 

\cleardoublepage
\bibliographystyle{IEEEtran}
\bibliography{mybib}
\end{document}